\documentclass[letterpaper, 10 pt, conference]{ieeeconf}  

\IEEEoverridecommandlockouts                              

\usepackage{graphicx}
\usepackage{url}  

\usepackage{amsmath}
\DeclareMathOperator*{\argmax}{argmax}

\title{\LARGE \bf
Deep Learning Framework Applied For Predicting Anomaly of Respiratory Sounds
}

\author{Dat~Ngo$^{1}$,
             Lam~Pham$^{1}$, 
             Anh~Nguyen$^{1}$, 
             Ben~Phan$^{2}$,
             Khoa~Tran$^{2}$,
             Truong~Nguyen$^{1}$ %
\thanks{D. Ngo, L. Pham, A. Nguyen, and T. Nguyen are with Electrical and Electronics Department, Ho Chi Minh City University of Technology, Vietnam.}%
\thanks{B. Phan, K. Tran are with Department of Electrical Engineering, Da Nang University of Science and Technology, Vietnam.}%
}

\begin{document}

\maketitle
\thispagestyle{empty}
\pagestyle{empty}

\begin{abstract}

This paper proposes a robust deep learning framework used for classifying anomaly of respiratory cycles. Initially, our framework starts with front-end feature extraction step. This step aims to  transform the respiratory input sound into a two-dimensional spectrogram where both spectral and temporal features are well presented. Next, an ensemble of C-DNN and Autoencoder networks is then applied to classify into four categories of respiratory anomaly cycles. In this work, we conducted experiments over 2017 Internal Conference on Biomedical Health Informatics (ICBHI) benchmark dataset. As a result, we achieve competitive performances with ICBHI average score of 0.49, ICBHI harmonic score of 0.42.

\indent \textit{Clinical relevance}--- Respiratory disease, wheeze, crackle, ensemble, C-DNN, autoencoder network 
\end{abstract}

\section{INTRODUCTION}

According to statistics of Global Burden of Diseases, Injuries and Risk Factors Study, there is an alarming number of deaths due to chronic respiratory diseases, with the figure increased from 3.32 million in 1990 to 3.91 million in 2017~\cite{li2020trends}. Furthermore, it becomes worse when this number is expected to continue go up in the next ten years. However, with the timely development of respiratory research, most respiratory diseases nowadays would be preventable by the early diagnosis. For instance, lung auscultation has been introduced as one of the most inexpensive, noninvasive and time-saving methods for respiratory examination thanks to every respiratory cycle can be heard and detected as whether its sound is normal or not. In particular, to better spread effective prevention and treatment widely for respiratory diseases, a reliable and quantitative diagnosis support method such as Computer-Aided Diagnosis (CAD) system~\cite{doi2007computer} is proposed. This systems is in an attempt of supporting doctors to hear, detect and differentiate automatically  between different respiratory sound patterns~\cite{kandaswamy2004neural}. Inspired from this, analysing respiratory sound by robust machine learning methods has recently attracted much attention. Particularly, authors in~\cite{lung_hmm_01} ultilized Mel-frequency cepstral coefficient (MFCC) as a frame-based feature representation to represent lung sounds into featuring vectors. Next, conventional machine learning models such as Hidden Markov Model~\cite{lung_hmm_01}, Support Vector Machine~\cite{lung_svm_01}, and Decision Tree~\cite{lung_tree_18} explored these vectors to classify anomalies of respiratory sounds. On the other hand, some researchers laid an emphasis on further analysis on feature extraction step via two-dimensional spectrogram. This is applied in order to fully represent audio features like an image in both temporal and spectral information, and then classified by more powerful architectures from image processing such as CNN~\cite{lung_cnn_01, lung_cnn_02} and  RNN~\cite{lung_rnn_01, lung_rnn_02}. Although many machine learning methods participated in this field, here is an inconsistency between dataset and performance comparison among publications. For instance, some authors in~\cite{mendes2016detection,datta2017automated,oletic2019hidden,shi2019lung,messner2018crackle} evaluated their systems over unpublished datasets. Furthermore, it is hard to compare performance when systems proposed use different ratio for splitting data, especially patient's objects.

To tackle these issues, we evaluate our proposed systems over the 2017 Internal Conference on Biomedical Health Informatics (ICBHI)~\cite{ic_dataset}, one of the largest dataset of respiratory sound published.
In terms of the system proposed, we approach deep learning based framework. In particular, we use Gammatone filter to generate Gamatonegram spectrogram where both spectral and temporal information are well represented. Next, the spectrogram is explored by an ensemble of C-DNN and Autoencoder networks.

\section{icbhi dataset and task defined}

\subsection{ICBHI dataset}
\label{icbhi}

The 2017 Internal Conference on Biomedical Health Informatics (ICBHI)~\cite{ic_dataset} is one of the largest annotated dataset of respiratory sounds published. Specifically, it contains 920 audio recordings collected in several years from 126 subjects in two different European countries. 
The subjects are identified as being healthy or exhibiting one of the following respiratory diseases or conditions such as: COPD, Bronchiectasis, Asthma, upper and lower respiratory tract infection, Pneumonia, Bronchiolitis. All recordings account for the duration of 5.5 hours, comprising 6898 respiratory cycles professionally labeled by respiratory experts. Within each audio recording, four different types of respiratory cycle are denoted as \textit{Crackle}, \textit{Wheeze}, \textit{Both} (\textit{Crackle \& Wheeze}), and \textit{Normal} according to the identified onset (i.e. starting time) and offset (i.e. ending time). Furthermore, these cycles show
 various duration ranging from 0.2\,s up to 16.2\,s and unbalanced (i.e. 1864 cycles of \textit{Crackle},  886 cycles of \textit{Wheeze}, 506 cycles of \textit{Both}, and 3642 cycels of \textit{Normal}).

\subsection{Task defined from ICBHI dataset}
\label{tasks}

\begin{table}[th]
    \caption{Confusion matrix of anomaly cycle classification.} 
        	\vspace{-0.2cm}
    \centering
    \begin{tabular}{l l l l l} 
        \hline 
                               &  \textbf{Crackle}  &  \textbf{Wheeze} &  \textbf{Both}  &  \textbf{Normal} \\
        \hline 
             \textbf{Crackle}  	& $C_{c}$            & $W_{c}$          & $B_{c}$         & $N_{c}$ \\        
             \textbf{Wheeze}   & $C_{w}$            & $W_{w}$          & $B_{w}$         & $N_{w}$ \\
             \textbf{Both}		& $C_{b}$            & $W_{b}$          & $B_{b}$         & $N_{b}$ \\
             \textbf{Normal}   & $C_{n}$            & $W_{n}$          & $B_{n}$         & $N_{n}$ \\
       \hline 
             \textbf{Total}    	& $C_{t}$            & $W_{t}$          & $B_{t}$         & $N_{t}$ \\
       \hline 
    \end{tabular}
    \label{table:cycle_tab} 
\end{table}

 Given by ICBHI dataset, this paper evaluates performance of respiratory anomaly classification among four different cycles (\textit{Crackle}, \textit{Wheeze}, \textit{Both}, and \textit{Normal}).
 In terms of metric used for evaluating, we follow ICBHI challenge, thus report ICBHI scores as mentioned in~\cite{ic_dataset}.
In particular, a confusion matrix of respiratory cycle classified  is presented in Table \ref{table:cycle_tab}. Specifically, the letters of \textit{C, W, B,} and \textit{N} denote the numbers of cycles of  \textit{Crackle}, \textit{Wheeze}, \textit{Both}, and \textit{Normal}, respectively, whereas \textit{c, w, b,} and \textit{n} subscripts indicate the inference results. The sums $C_{t}$, $W_{t}$, $B_{t}$ and $N_{t}$ are the total numbers of cycles. Thus, \textit{Sensitivity (SE)} , and  \textit{Specitivity (SP)} are firstly computed by, 

 \begin{equation}
     \label{eq:task_1_1_sen}
     Sensitivity =  \frac{C_{c} + W_{w} + B_{b}}{C_{t} + W_{t} + B_{t}}
 \end{equation}

 \begin{equation}
Specificity =  \frac{N_{n}}{N_{t}}
 \end{equation}
%
Next, ICBHI scores comprising average score (AS) and the  harmonic score (HS) are compuated by,

\begin{equation}
AS =  \frac{SE+SP}{2}
 \end{equation}

\begin{equation}
HS =  \frac{2.SE.SP}{SE+SP}
 \end{equation}

\section{Deep learning based framework proposed}
The proposed high-level system architecture including two main parts: front-end feature extraction as described in the upper part of Fig. 1 with setting parameters in Table II and back-end deep learning model as shown in the lower part of Fig. 1. 

\subsection{Front-end feature extraction}

In particular, we re-sample respiratory cycles to 4000 Hz since frequency banks of abnormal sounds (\textit{Crackle} and \textit{Wheeze}) locate mostly from 60 to 2000 Hz. Consequently, re-sampled respiratory cycles showing different lengths are next duplicated to ensure the same length of 10 seconds. Next, respiratory cycles go through a bandpass filter of 100-2000 Hz to reduce noise. After that, these respiratory sounds are transformed into two-dimensional spectrograms by using Gammatone transformation. To generate Gammatone spectrogram (Gamma), we firstly compute Short-Time Fourier Transform (STFT) as presented below:

\textbf{Short-Time Fourier Transform (STFT):} The STFT spectrogram applies Fourier Transform to extract Frequency content of local section of input signal over short time duration.
Let consider $\mathbf{s}$($n$) as digital audio signal with length of \(N\) , a pixel value at central frequency \(f\) and time frame \(t\) of STFT spectrogram \(\mathbf{STFT}[F,T]\) is computed as:
\begin{equation}
\label{eq:stft}
\mathbf{STFT}[f,t] = \sum_{n=0}^{N-1} \mathbf{s}[n].\mathbf{w}[n]e^{-j2{\pi}fn}  
\end{equation}
where $\mathbf{w}$[$n$] is a window function, typically Hamming.
While time resolution ($T$) of STFT spectrogram is set by window side and hope size, the frequency resolution ($F$) equals to the number of central frequencies set to 1024. 
Then, we apply Gammatone filter into STFT spectrogram as described below:
%
\begin{figure}[t]
    \centering
    \vspace{-0.9cm}
    \includegraphics[width =0.9\linewidth]{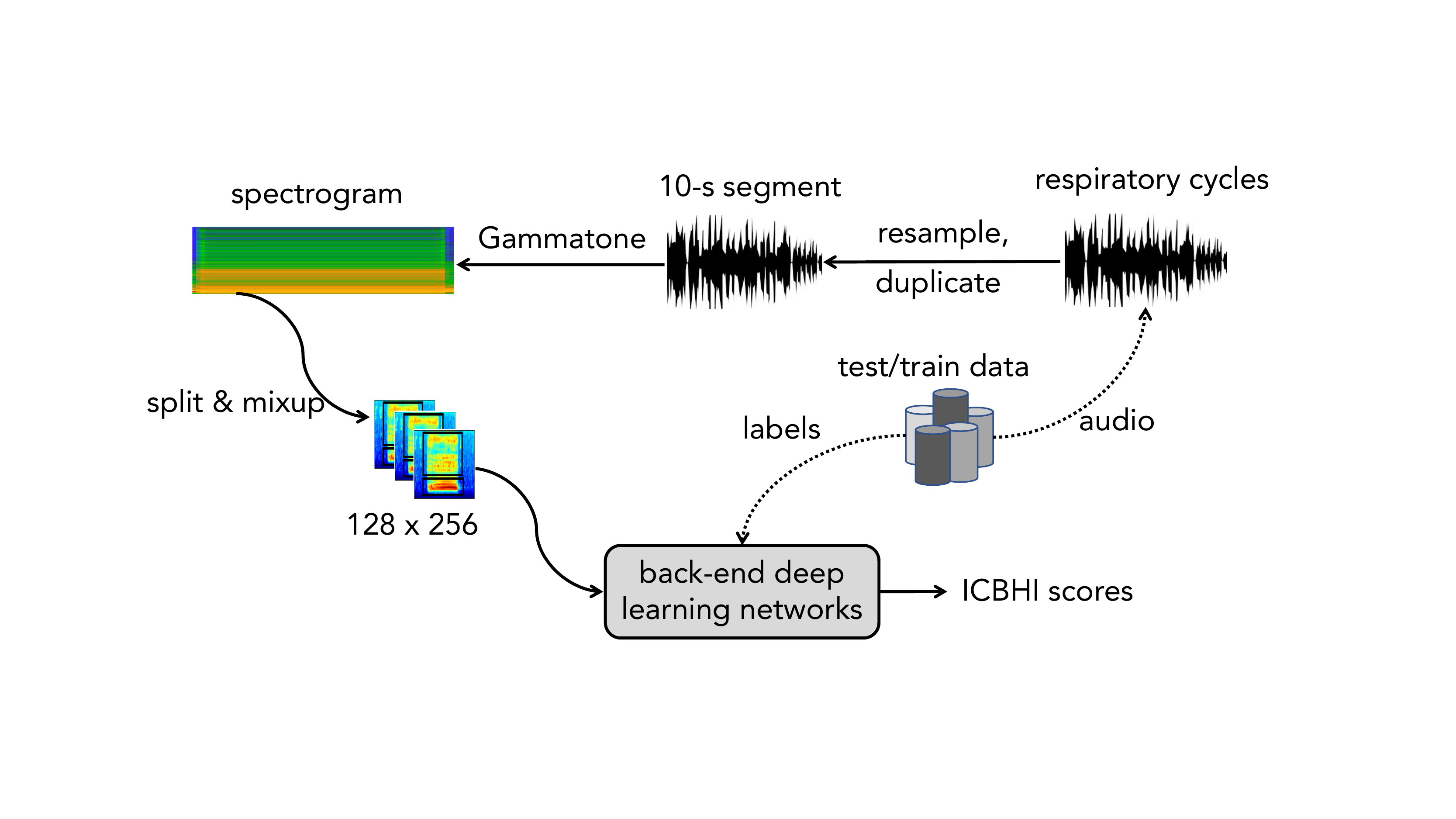}
    	\vspace{-0.9cm}
	\caption{Deep learning based framework proposed}
    \label{fig:A1}
\end{figure}
\begin{table}[t]
    \caption{Feature Extraction Parameter Setting} 
        	\vspace{-0.2cm}
    \centering
    \scalebox{1}{
    \begin{tabular}{l c} 
        \hline 
            \textbf{Factors}   &  \textbf{Setting}  \\
        \hline 
             Re-sample  & 4000 Hz \\         
             Cycle duration  & 10 s \\         
             Spectrogram & Gammatonegram~\cite{patterson1986auditory} \\         
             FFT number & 1024 \\   
             Window size & 0.2s \\ 
             Hop size & 0.04s \\              
             Patch size & $128\times256$ \\         
             Data augmentation & Random oversampling \& Mixup \\         
      \hline 
    \end{tabular}
    }
    \label{table:baseline} 
\end{table}
%

\textbf{Gammatone (GAM):} Gammatone filters are designed to model the frequency-selective cochlea activation response of the human inner ear~\cite{patterson1986auditory}, in which filter output simulates the frequency response of the basilar membrane.
The impulse response is given by:
\begin{equation}
\label{eq:gammaton}
g[k] = k^{P-1}T^{P-1}e^{-2bkT\pi}cos(2fkT\pi + \theta)    
\end{equation}
where \(k\) is time, \(P\) is the filter order, \(T\) is sampling period, \(b\) is filter bandwidth, \(f\) is central frequency, \(\theta\) is the phase of  the carrier.
The filter bank was then formulated as ERB scale~\cite{glasberg1990derivation} as:
\begin{equation}
\label{eq:gammaton}
ERB = 24.7(4.37.10^{-3}f + 1)
\end{equation}
To quickly generate Gamma spectrogram, we apply a toolbox developed by Ellis \emph{et al.}~\cite{ellis2009gammatone}, namely Gammatone-like spectrogram.
Firstly, audio signal is transformed into STFT spectra recently mentioned above.
Next, gammatone weighting $\mathbf{COE}[F_{gam}, F]$ is applied on STFT to obtain the Gamma spectrogram.
\begin{equation}
\label{eq:log-Mel}
\mathbf{GAM}[F_{gam},T] = \mathbf{COE}[F_{gam}, F]\times\mathbf{STFT}[F, T]
\end{equation}
where \(F_{gam}\) resolution of GAM spectrogram is Gammatone filter number of 128.

Next, each 10-s spectrogram of one respiratory cycle is thus split into non-overlapped patches of $128{\times}256$, likely an image. 
To deal with unbalanced data issue, we apply two data augmentation techniques on the image patches of $128{\times}256$. Firstly, we randomly oversample image patches to make sure that the number of patches per category is equal. Next, the mixup data augmentation \cite{salamon2017deep} is applied to enlarge Fisher’s criterion (i.e. the ratio of the between- class distance to the within class variance in the feature space) to increase variation of training data.
Let consider two original image patches as \(\mathbf{X_{1}}\), \(\mathbf{X_{2}}\) and expected labels as \(\mathbf{y_{1}}\), \(\mathbf{y_{2}}\), new image patches are generated as below equations:
\begin{equation}
\label{eq:mix_up_x1}
\mathbf{X_{mp1}} = \mathbf{X_{1}}\gamma + \mathbf{X_{2}}(1-\gamma) 
\end{equation}
\begin{equation}
\label{eq:mix_up_x2}
\mathbf{X_{mp2}} = \mathbf{X_{1}}(1-\gamma) + \mathbf{X_{2}}\gamma
\end{equation}
\begin{equation}
\label{eq:mix_up_y}
\mathbf{y_{mp1}} = \mathbf{y_{1}}\gamma + \mathbf{y_{2}}(1-\gamma)
\end{equation}
\begin{equation}
\label{eq:mix_up_y}
\mathbf{y_{mp2}} = \mathbf{y_{1}}(1-\gamma) + \mathbf{y_{2}}\gamma
\end{equation}
where \(\gamma\) is random coefficient from \textit{Beta} distribution, \(\mathbf{X_{mp1}}\), \(\mathbf{X_{mp2}}\) and \(\mathbf{y_{mp1}}\), \(\mathbf{y_{mp2}}\) are new image patches and labels generated, respectively.
Eventually, the mixup patches are fed into a back-end classifier, report the classification accuracy.

\subsection{Back-end classification}

\begin{figure*}[th]
    \centering
    \includegraphics[width =0.9\linewidth]{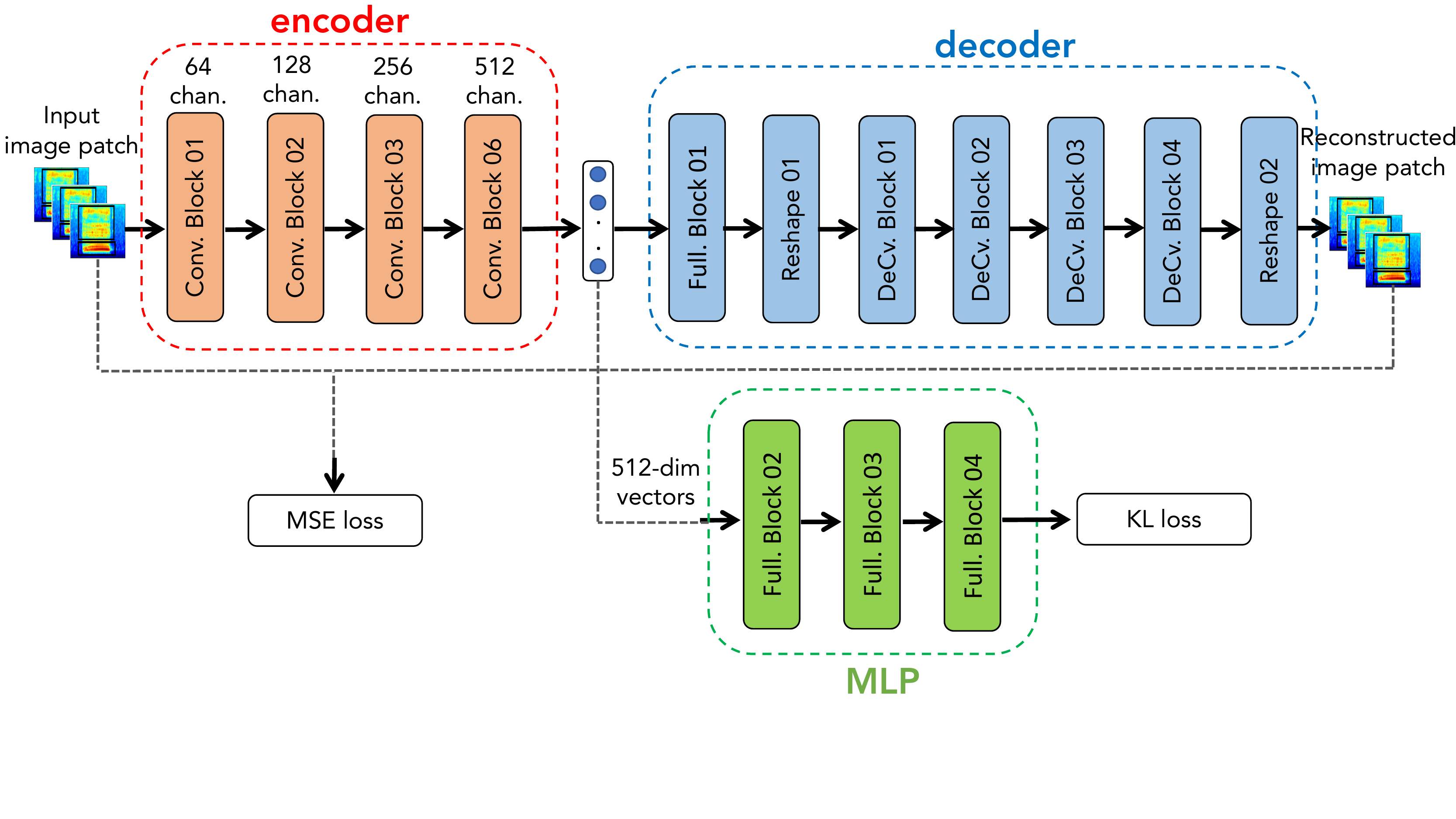}
    	\vspace{-1.4 cm}
    	\caption{Block-level architecture of Autoencoder network.}
    \label{fig:framework}
\end{figure*}
\begin{table}[h]
    \caption{C-DNN network architecture} 
        	\vspace{-0.2cm}
    \centering
    \scalebox{0.9}{
    \begin{tabular}{l c} 
        \hline 
            \textbf{Network architecture}   &  \textbf{Output}  \\
        \hline 
         \textbf{CNN} & \\
         Input layer (image patch of $128{\times}256$)  &        \\
         Bn - Cv [$3{\times}3$] - Relu - Bn - Mp [$2{\times}2$] - Dr ($10\%$)      & $62{\times}78{\times}64$\\
         Bn - Cv [$3{\times}3$] - Relu - Bn - Mp [$2{\times}2$] - Dr ($15\%$)      & $31{\times}39{\times}128$\\
         Bn - Cv [$3{\times}3$] - Relu - Bn - Mp [$2{\times}2$] - Dr ($20\%$)      & $16{\times}20{\times}256$\\
         Bn - Cv [$3{\times}3$] - Relu - Bn - Gmp - Dr ($25\%$)       & $512$\\
         \hline
         \textbf{DNN} & \\
         Input layer ($512$-dimensional vectors)  &        \\
         Fl - Relu - Dr ($30\%$) &  $1024$       \\
         Fl - Softmax & 4 \\
       \hline 
    \end{tabular}
    }
    \label{table:CDNN} 
\end{table}
%
For back-end classification, we propose an ensemble of C-DNN and autoencoder networks in this paper.
As regards C-DNN network architecture, it comprises two main parts of CNN and DNN as shown in Table \ref{table:CDNN}, likely Lenet-6~\cite{perna2018convolutional}. The CNN as the upper part in Table \ref{table:CDNN} performs batch normalization (Bn), convolutional (Cv[kernel size]), rectified linear units (Relu), max pooling (Mp[kernel size]) and global max pooling (Gmp), and dropout (Dr (dropout percentage)) layers. Meanwhile, the DNN as shown in the lower part in Table \ref{table:CDNN} comprises two fully-connected (Fl), rectified linear units (Relu) and a final Softmax layer for classification.

In terms of autoencoder network architecture proposed, it shows more complicated with two training phases as showed in Fig. \ref{fig:framework}. 
At the first phase, we present an Encoder-Decoder architecture which is used to extract embedding vectors containing condensed information. 
Next, the embeddings are fed into a MLP based network architecture for classifying into four categories in the second phase.
As shown in Table \ref{table:auto} and the upper part of Fig. \ref{fig:framework}, Encoder part of Encoder-Decoder architecture comprises four Conv. Blocks each which performs layers as same as C-DNN network architecture. 
The Encoder helps to compress input image patch into condensed vectors, referred to as embeddings.
Meanwhile, Decoder firstly use a fully-connected layer to decompress embeddings, thus apply four DeCv. Blocks (i.e. each DeCv. Block comprises a de-convolutional (DeCv[kernel size]) layer and a rectified linear unit layer (Relu)) to re-construct the input image patch). 
Notably, batch normalization (Bn), max pooling (Mp[kernel size]) and global max pooling (Gmp) and dropout (Dr (dropout percentage)) layers are not applied in Decoder.
For MLP-based network architecture as shown in Table \ref{table:MLP} and the lower part of Fig. \ref{fig:framework}, it is configured by two fully-connected layers (Fl). The first fully-connected layer follow by a ReLu and a Dr. Meanwhile, a Softmax layer is used after the second fully-connected layer for classification.

\begin{table}[t]
    \caption{Encoder-decoder network architecture} 
        	\vspace{-0.2cm}
    \centering
    \scalebox{0.8}{
    \begin{tabular}{l l c} 
        \hline 
        \textbf{Block} &\textbf{Network architecture}   &  \textbf{Output}  \\
        \hline 
          &\textbf{Encoder} & \\
         &Input layer (image patch of $128{\times}256$)  &        \\
         Conv. Block 01 &Bn - Cv [$3{\times}3$] - Relu - Bn - Mp [$2{\times}2$] - Dr ($10\%$)      & $64{\times}128{\times}64$\\
         Conv. Block 02 &Bn - Cv [$3{\times}3$] - Relu - Bn - Mp [$2{\times}2$] - Dr ($15\%$)      & $32{\times}64{\times}128$\\
         Conv. Block 03 &Bn - Cv [$3{\times}3$] - Relu - Bn - Mp [$2{\times}2$] - Dr ($20\%$)      & $16{\times}32{\times}256$\\
         Conv. Block 04 &Bn - Cv [$3{\times}3$] - Relu - Bn - Gmp - Dr ($25\%$)       & $512$\\
         \hline
         &\textbf{Decoder} & \\
         &Input layer ($512$-dimensional vectors)  &        \\
         Full. Block 01 &Fl - Relu &  $32768$       \\
         Reshape 01 &Reshape &  $8{\times}16{\times}256$ \\
         DeCv. Block 01 &DeCv [$3{\times}3$] - Relu &  $16{\times}32{\times}128$ \\
         DeCv. Block 02 &DeCv [$3{\times}3$] - Relu &  $32{\times}64{\times}64$ \\
         DeCv. Block 03 &DeCv [$3{\times}3$] - Relu &  $64{\times}128{\times}32$ \\
         DeCv. Block 04 &DeCv [$3{\times}3$] - Relu &  $128{\times}256{\times}1$ \\
         Reshape 02 &Reshape &  $128{\times}256$ \\
       \hline 
    \end{tabular}
    }
    \label{table:auto} 
\end{table}

\begin{table}[t]
    \caption{MLP-based network architecture} 
        	\vspace{-0.2cm}
    \centering
    \scalebox{0.9}{
    \begin{tabular}{l l c} 
        \hline 
            \textbf{Block}&\textbf{Network architecture}   &  \textbf{Output}  \\
        \hline 
        &Input layer ($512$-dimensional vectors)  &        \\
         Full. Block 02&Fl - Relu - Dr ($50\%$) &  $1024$       \\
         Full. Block 03&Fl - Relu - Dr ($50\%$) &  $1024$       \\
         Full. Block 04&Fl - Softmax & 4 \\
       \hline 
    \end{tabular}
    }
    \label{table:MLP} 
\end{table}

\subsection{Experimental setting}
\begin{table}[ht]
    \caption{ICBHI dataset splitting} 
        	\vspace{-0.2cm}
    \centering
    \begin{tabular}{l c c} 
        \hline 
                               &  \textbf{Training Set}  &  \textbf{Test Set}  \\
        \hline 
              \textit{Wheezes}   & 501           & 385         \\
             \textit{Crackles}		& 1215            & 649          \\
             \textit{Both}    	& 363            & 143          \\
             \textit{Normal}    	& 2063          & 1579         \\
       \hline 
    \end{tabular}
    \label{table:splitting detail} 
\end{table}
Given by ICBHI dataset, we follow ICBHI challenge setting, thus divide into Training and Test subsets with the ratio of 60\% and 40\% respectively as shown in Table \ref{table:splitting detail}. Notably, the splitting proposed prevents the present of object's audio recordings on both Training and Test subsets.

In terms for back-end network architectures proposed, we adopt Tensorflow framework and set learning rate to 0.0001, a batch size of 50, epoch number  of 100, and Adam method \cite{kingma2014adam} for learning rate optimization. 
As using mixup data augmentation, the labels are not one-hot format.
Therefore, we use Kullback-Leibler (KL) divergence loss \cite{kullback1951information} in C-DNN and MLP-based networks instead of the standard cross-entropy loss as shown in Eq. (\ref{eq:kl_loss}) below:

\begin{align}
\label{eq:kl_loss}
Loss_{KL}(\Theta) = \sum_{n=1}^{N}\mathbf{y}_{n}\log(\frac{\mathbf{y}_{n}}{\mathbf{\hat{y}}_{n}})  +  \frac{\lambda}{2}||\Theta||_{2}^{2},
\end{align}
where \(Loss_{KL}(\Theta)\) is KL-loss function, $\Theta$ describes the trainable parameters of the network trained, $\lambda$ denote the $\ell_2$-norm regularization coefficient experimentally set to 0.0001, \(N\) is the batch size,
$\mathbf{y_{n}}$ and $\mathbf{\hat{y}_{n}}$  are the ground-truth and the network recognized output, respectively.
To train Encoder-Decoder network, we use Mean Squared Error (MSE) loss to compare original image patches (input of Encoder) to reconstructed image patches (output of Decoder) as below,

\begin{align}
\label{eq:kl_loss}
Loss_{MSE} = \frac{1}{2N}\sum_{n=1}^{N}(\mathbf{X}_{n}-\mathbf{\hat{X}}_{n})^{2}
\end{align}
where $\mathbf{X}_{n}$ and $\hat{\mathbf{X}}_{n}$ are input image patch and reconstructed image patch, respectively.

\subsection{Late fusion strategy}
As the C-DNN model work on patch level, the probability of an entire spectrogram is computed by averaging of all patches' probabilities. Let consider $\mathbf{P_{C-DNN}^{n}} = (\mathbf{k_{1}^{n}, k_{2}^{n},..., k_{C}^{n}})$,  with $C$ being the category number and the \(n^{th}\) out of \(N\) patches fed into learning model, as the probability of a test sound instance, then the mean classification probability is denoted as  \(\mathbf{\bar{p}_{C-DNN}} = (\bar{k}_{1}, \bar{k}_{2}, ..., \bar{k}_{C})\) where,

\begin{equation}
    \label{eq:mean_stratergy_patch}
    \bar{k}_{c} = \frac{1}{N}\sum_{n=1}^{N}k_{c}^{n}  ~~~  for  ~~ 1 \leq c \leq C 
\end{equation}

As the Autoencoder model work on patch level, the probability of an entire spectrogram is computed by averaging of all patches' probabilities. Let consider $\mathbf{P_{MLP}^{n}} = (\mathbf{m_{1}^{n}, m_{2}^{n},..., m_{C}^{n}})$,  with $C$ being the category number and the \(n^{th}\) out of \(N\) patches fed into MLP-based network, as the probability of a test sound instance, then the mean classification probability is denoted as  \(\mathbf{\bar{p}_{MLP}} = (\bar{m}_{1}, \bar{m}_{2}, ..., \bar{m}_{C})\) where,

\begin{equation}
\label{eq:mean_stratergy_patch}
\bar{m}_{c} = \frac{1}{N}\sum_{n=1}^{N}m_{c}^{n}  ~~~  for  ~~ 1 \leq c \leq C 
\end{equation}

To evaluate the ensemble of C-DNN and Autoencoder, we propose three late fusion schemes, namely \textit{Max}, \textit{Mean}, and \textit{Mul} fusions.
 
The probability of combination with \textit{Max} strategy \(\mathbf{p_{f-max}}\) is obtained by,

\begin{equation}
\label{eq:mix_up_x1}
\mathbf{p_{f-max}} = max(\mathbf{\bar{p}_{C-DNN}} , \mathbf{\bar{p}_{MLP}})
\end{equation}

The probability of combination with \textit{Mean} strategy \(\mathbf{p_{f-mean}}\) is obtained by,

\begin{equation}
\label{eq:mix_up_x1}
\mathbf{p_{f-mean}} = \frac{\mathbf{\bar{p}_{C-DNN}} + \mathbf{\bar{p}_{MLP}}}{2}        
\end{equation}

The probability of combination with \textit{Mul} strategy \(\mathbf{p_{f-mul}}\) is obtained by,

\begin{equation}
\label{eq:mix_up_x1}
\mathbf{p_{f-mul}} = \frac{\mathbf{\bar{p}_{C-DNN}}  .  \mathbf{\bar{p}_{MLP}}}{2}        
\end{equation}

Eventually, the predicted result is decided by,

\begin{equation}
\label{eq:average}
\hat{y} = \argmax_{c \in \{1,2,\ldots,C\}}\bar{p}_c.
\end{equation}
\section{Experimental results and discussion}

As details shown in Table \ref{table:results}, it can be seen that Autoencoder is better than C-DNN in terms of SE score, improving by 0.02.
By contrast, the SP score of Auto-encoder reduces by 0.01 compared to C-DNN.
As a result, both networks achieve the same AS score of 0.47. Meanwhile, HS scores present 0.41 and 0.43 for C-DNN and Autoencoder, respectively. 

As regards comparison among three late fusion methods, Mean fusion achieves the highest performances with SP score of 0.69, SE score of 0.30, and AS/HS scores of 0.49/0.42. 
Compared to individual C-DNN or Autoencoder model, although ensemble methods make SE scores reduce a little, they help to improve SP scores significantly.

Compare to the state-of-the-art systems as shown in Table \ref{table:comp} (note that we only compare to systems which follow splitting ratio of 60/40 defined by ICBHI challenge), while our system's SP ranks fourth position, SE score achieves the top three.  
In terms of only using single spectrogram, our system achieves very competitive AS/HS scores of 0.49/0.42 that is top two after the systems proposed in~\cite{ic_cnn_19_iccas}.
\begin{table}[t]
	\caption{Performance of C-DNN, Autoencoder, and their fusions (highest scores in \textbf{bold})} 
        	\vspace{-0.2cm}
    \centering
    \scalebox{0.9}{
    \begin{tabular}{l c c c c c c} 
        \hline 
	   \textbf{Systems}    &\textbf{SP}  &\textbf{SE}   &\textbf{AS/HS Scores}  \\
        \hline 
	C-DNN            &0.63             &0.31            &0.47/0.41  \\
	Autoencoder    &0.62             &0.33            &0.47/0.43  \\
	    Max fusion      &0.67            &0.30             &0.48/0.42  \\
	    Mean fusion      &\textbf{0.69}            &\textbf{0.30}             &\textbf{0.49/0.42}  \\
	    Mul fusion      &\textbf{0.69}            &0.29             &0.49/0.41\\             
         \hline 
    \end{tabular}
    }
    \label{table:results} 
\end{table}
%
\begin{table}[t]
    \caption{Compare our systems (the lower part) against state-of-the-art systems with ICBHI challenge splitting (highest scores in \textbf{bold}).} 
        	\vspace{-0.2cm}
    \centering
    \scalebox{0.9}{

    \begin{tabular}{l l l l c c c} 
        \hline 
	    \textbf{Features}    & \textbf{Classifiers}                  &\textbf{SP}   &\textbf{SE}   &\textbf{AS/HS Scores}  \\
        \hline 
       MFCC&Decision Tree~\cite{ic_baseline}                    &0.75             &0.12           &0.43/0.15  \\        
        MFCC&HMM~\cite{ic_hmm_18_sp}                   &0.38             &\textbf{0.41}           &0.39/0.23  \\        
       STFT+Wavelet&SVM~\cite{ic_svm_18_sp}                   &0.78             &0.20           &0.47/0.24  \\
        Gammatonegram &CNN-MoE          &0.68 &0.26 & 0.47/0.37   \\ 
        log-Mel&CNN-RNN~\cite{ic_cnn_19_iccas}          &0.69 &0.30 & 0.50/\textbf{0.46}    \\    
       Scalogram&CNN-RNN~\cite{ic_cnn_19_iccas}              &0.62 &0.37 & 0.50/\textbf{0.46}    \\     
       log-Mel+Scalogram & CNN-RNN~\cite{ic_cnn_19_iccas}              &\textbf{0.81} &0.28 & \textbf{0.54}/0.42    \\       
               \hline 
       \textbf{Gammatonegram}     &   \textbf{C-DNN}                        &0.63    &0.31   & 0.47/0.41    \\
        \textbf{Gammatonegram}     &   \textbf{Autoencoder}                        &0.62    &0.33   & 0.47/0.43    \\
       \textbf{Gammatonegram}     &   \textbf{C-DNN+Autoencoder}                        &0.69    &0.30   & 0.49/0.42    \\

       \hline 
    \end{tabular}
    }
    \label{table:comp} 
\end{table}

\section{Conclusion}

We have just presented a deep learning based framework which is used for classifying respiratory sound. The exploration of Gammatone transformation and an ensemble of C-DNN and Autoencoder networks achieves significant performances of 0.49 and 0.42 in terms of ICBHI average and harmmonic scores over ICBHI benchmark dataset that are very competitive to the state-of-the-art systems.

\addtolength{\textheight}{-12cm}   

\bibliographystyle{IEEEbib}
\bibliography{refs}

\end{document}